\documentclass{article}
\usepackage{spconf,amsmath,graphicx,float}
\usepackage{multirow}
\usepackage{lscape}
\usepackage{booktabs}
\usepackage{nicematrix}
\usepackage{subcaption}
\usepackage{makecell}
\usepackage{amsmath}
\usepackage{amssymb}
\usepackage[normalem]{ulem}
\useunder{\uline}{\ul}{}
\usepackage{comment}
\usepackage{hyperref}

\title{Continual Learning for Out-of-Distribution Pedestrian Detection}

\name{Mahdiyar Molahasani, Ali Etemad, Michael Greenspan}
\address{Dept. ECE \& Ingenuity Labs Research Institute, Queen's University, Kingston, Canada}
\begin{document}

\maketitle

\begin{abstract}
A continual learning solution is proposed to address the out-of-distribution generalization problem for pedestrian detection. While recent pedestrian detection models have achieved impressive performance on various datasets, they remain sensitive to shifts in the distribution of the inference data. 
Our method adopts and modifies Elastic Weight Consolidation to a backbone object detection network, in order to penalize the changes in the model weights based on their importance towards the initially learned task. We show that when trained with one dataset and fine-tuned on another, our solution learns the new distribution and maintains its performance on the previous one, avoiding catastrophic forgetting. We use two popular datasets, CrowdHuman and CityPersons for our cross-dataset experiments, and show considerable improvements over standard fine-tuning,
with a $9\%$ and $18\%$ miss rate percent reduction improvement
in the CrowdHuman and CityPersons datasets, respectively.
\end{abstract}


\section{Introduction}
\label{sec:intro}

Pedestrian detection is crucial to many machine vision applications such as autonomous vehicles, surveillance systems, and behavior understanding~\cite{hasan2021generalizable}. Recently, powerful deep learning models using Convolutional Neural Networks (CNNs) have achieved strong performance in pedestrian detection. However, despite steady improvements in performance, many existing solutions suffer performance degradation in cross-dataset scenarios~\cite{hasan2021generalizable}. This occurs when the distributions of the train and test datasets differ, which is known as the \textit{out-of-distribution generalization} problem \cite{Lin2022OnCM}.
This occurs, for example, in pedestrian detection applications when the train and test datasets are acquired in different weather conditions or cities.


One solution that has been proposed is to fine-tune a trained model on a new dataset, which can allow the model to adapt to the new data distribution. However, this approach can lead to \textit{catastrophic forgetting}, wherein the model's performance on the initial distribution degrades significantly~\cite{kirkpatrick2017overcoming}.
As a response, 
\textit{Continual Learning} (CL) has been proposed as a collection of techniques designed specifically to overcome catastrophic forgetting~\cite{de2021continual}. 
While continual learning remains an active area of research, most advances have been focused on classification tasks, and to our knowledge there have not been any applications of continual learning to the regression tasks inherent in out-of-distribution pedestrian detection.

In this paper, we address the out-of-distribution generalization problem in pedestrian detection using continual learning. We modify the Elastic Weight Consolidation (EWC)~\cite{kirkpatrick2017overcoming} loss and combine it with Faster R-CNN~\cite{zhang2016faster} to create a 
solution that exhibits robust continual learning. Our proposed method allows the model to learn the new distribution without forgetting the previous one, by leveraging the parameters that are less important to the previous distribution. We evaluate the performance of our method on the CityPersons and CrowdHuman datasets in a cross-dataset setup. Our experiments show that the model can effectively learn the two distributions together, with few negative side-effects.



The contributions of this work are: 
(\textbf{1}) We propose a novel continual learning-based solution to introduce out-of-distribution generalization for pedestrian detection. Our solution exhibits robust performance on two popular pedestrian detection datasets used in a cross-dataset setup.
(\textbf{2}) We adapt EWC to allow its application to the regression task inherent in pedestrian detection.
(\textbf{3}) Upon acceptance, we will make our code public to contribute to the field.


\begin{figure*}[t]
\begin{center}
\includegraphics[width=0.95\linewidth ]{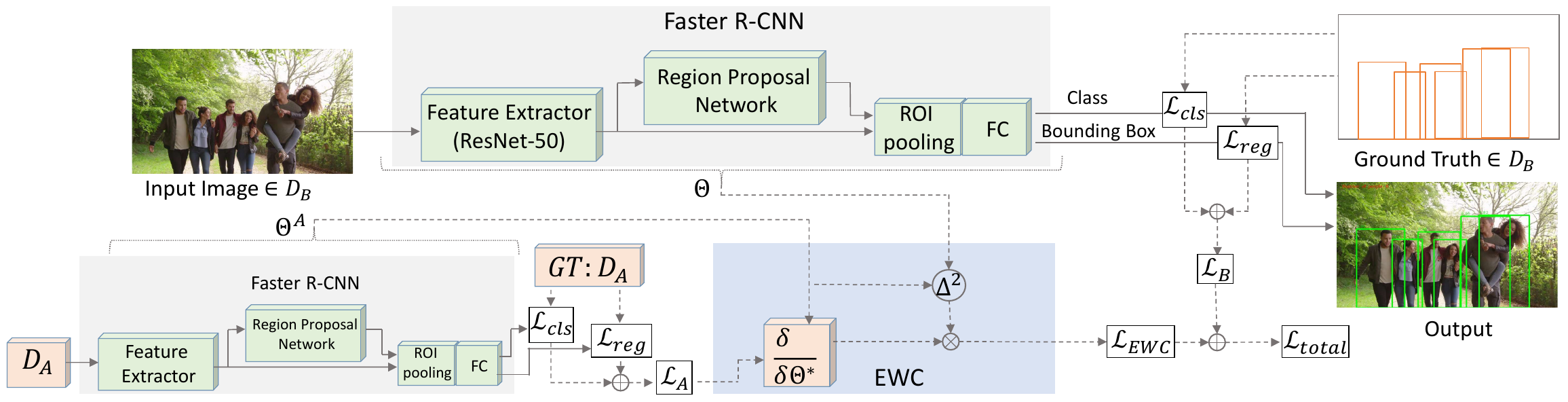}
\end{center}
\vspace{-0.5 cm}
   \caption{The architecture of the proposed model. Dashed lines are used solely in the training stage.}
\label{arch}
\end{figure*}

\section{Related Work}
\noindent \textbf{Pedestrian Detection.}
In recent years, CNNs have become the most employed solution for pedestrian-related applications  \cite{campmany2016gpu,zand2022multiscale,hbaieb2019pedestrian}.
One of the CNNs designed specifically for object detection is R-CNN \cite{girshick2014rich}.
Its promising results have therefore made it a quite popular backbone for pedestrian detection~\cite{hosang2015taking,zhang2016far}. Faster R-CNN was subsequently proposed to enhance the performance and the frame rate of previous models~\cite{zhang2016faster}. Another variation of R-CNN is Cascade R-CNN~\cite{cai2019cascade}. This architecture has multiple detection heads and removes more difficult false positive predictions, progressively. Although these models achieve strong results in pedestrian detection on various datasets, they are vulnerable to cross-dataset testing protocols~\cite{hasan2021generalizable}, which can often be attributed to distribution shift between the datasets. 

\noindent \textbf{Out-of-Distribution Learning.}
Out-of-distribution generalization aims to address the scenario when the distribution of the test data is not the same as the distribution of the training set. 
The existing methods in this area can be categorized based on the way they tackle the problem as: unsupervised representation learning, supervised model learning, and optimization for out-of-distribution generalization~\cite{shen2021towards}. 
The purpose of these approaches is to make the models more robust to the shift in the distribution of the training data~\cite{ganin2016domain,peng2018synthetic}.

\noindent \textbf{Continual Learning.} 
Continual learning algorithms enable networks to learn a new task without forgetting the previous one. These methods have been employed for tackling the out-of-distribution generalization problem in prior works ~\cite{lin2022continual}.
Elastic Weight Consolidation (EWC) is a popular regularization-based continual learning method~\cite{kirkpatrick2017overcoming}, which penalizes the changes in model weights based on their importance. This importance is calculated based on the Fisher Information Matrix. Synaptic Intelligence is another technique, which unlike EWC, calculates the importance in real-time in an online manner~\cite{zenke2017continual}. Another approach, RWalk, tackles this problem from a KL-divergence-based perspective~\cite{chaudhry2018riemannian}. 

Here we rely on a modified version of EWC to leverage continual learning to perform out-of-distribution pedestrian detection. To our knowledge, continual learning has not been explored in prior work for 
pedestrian detection 
in an \textit{out-of-distribution} scenario. 
This is likely due to the fact that standard EWC and its descendants are used for either classification or regression individually, but not for both tasks simultaneously,
as is required for pedestrian detection.



\begin{table*}[t!]
\footnotesize
\centering
\caption{Miss rate ($MR^{-2}(\downarrow)$) of two test scenarios  for various occlusion ratios. }

\begin{tabular}{c|c|cccc|cccc|cccc}
\Xcline{3-14}{1\arrayrulewidth}
\noalign{\vskip 0.7mm}
\multicolumn{2}{c}{} &
\multicolumn{4}{c|}{\textbf{Train}}                               & \multicolumn{4}{c|}{\textbf{Baseline Fine-tune}}      & \multicolumn{4}{c}{\textbf{Proposed}}                        \\ \cline{1-14} 
\multirow{2}{*}{\textbf{Scenario}} & \multirow{2}{*}{\textbf{Test Dataset}} &
\multicolumn{4}{c|}{CrowdHuman}                                   & \multicolumn{4}{c|}{CityPersons}                      & \multicolumn{4}{c}{CityPersons}                                   \\ \cline{3-14} 
                                   &                                        & Reas.          & Bare           & Partial        & Heavy          & Reas.       & Bare        & Partial     & Heavy       & Reas.          & Bare           & Partial        & Heavy          \\ \noalign{\vskip 0.7mm}
                                   \hline
                                   \noalign{\vskip 0.7mm}
\multirow{2}{*}{1}                 & CityPersons                            & 32.65          & 24.66          & 36.28          & 76.04          & {\ul 25.28} & {\ul 17.69} & {\ul 28.42} & {\ul 66.44} & \textbf{23.68} & \textbf{16.67} & \textbf{25.63} & \textbf{65.65} \\
                                   & CrowdHuman                             & \textbf{55.48} & \textbf{40.80} & \textbf{48.62} & \textbf{73.07} & 63.20       & 50.64       & 63.93       & 85.33       & {\ul 58.49}    & {\ul 45.30}    & {\ul 57.44}    & {\ul 81.18}    \\           \noalign{\vskip 0.7mm} 
                       \hline
                       \hline
                       \noalign{\vskip 0.7mm}
\multirow{2}{*}{\textbf{Scenario}} & \multirow{2}{*}{\textbf{Test Dataset}} & \multicolumn{4}{c|}{CityPersons}                                  & \multicolumn{4}{c|}{CrowdHuman}                       & \multicolumn{4}{c}{CrowdHuman}                                    \\ \cline{3-14} 
                                   &                                        & Reas.          & Bare           & Partial        & Heavy          & Reas.       & Bare        & Partial     & Heavy       & Reas.          & Bare           & Partial        & Heavy          \\ \noalign{\vskip 0.7mm}
                                   \hline
                                   \noalign{\vskip 0.7mm}
\multirow{2}{*}{2}                 & CityPersons                            & \textbf{23.39} & \textbf{15.01} & \textbf{25.86} & \textbf{67.39} & 35.96       & 27.02       & 39.90       & 78.72       & {\ul 31.42}    & {\ul 23.56}    & {\ul 36.39}    & {\ul 74.39}    \\
                                   & CrowdHuman                             & 64.23          & 47.80          & 67.59          & 91.56          & {\ul 58.96} & {\ul 43.91} & {\ul 56.80} & {\ul 82.20} & \textbf{55.97} & \textbf{41.31} & \textbf{48.74} & \textbf{73.62} \\ \noalign{\vskip 0.7mm}

\hline
                                   
\end{tabular}
\label{res_i}
\end{table*}

\section{Proposed Method}
Let the pedestrian detection problem be denoted as:
\begin{equation}\label{eq:1}
  \begin{array}{l}
Dets^A = \mathcal{P}(\theta^A,\Lambda,\mathcal{D}^A_{t})\\
=\{cls(\theta^A,\Lambda,\mathcal{D}^A_{t}),reg(\theta^A,\Lambda,\mathcal{D}^A_{t})\}
 \end{array}
\end{equation}
Here $\mathcal{P}(\cdot)$ performs pedestrian detection on three inputs, where $\mathcal{D}_{t}^A$ is either training or testing partition of dataset $\mathcal{D}^A$ depending on the context, $\theta^A$ are the model parameters after training on $\mathcal{D}_{Train}^A$, and $\Lambda$ is a set of hyperparameters including anchor box specifications. 
$\mathcal{P}(\cdot)$ consists of $cls(\cdot)$ and $reg(\cdot)$ which classifies the objects and estimates their corresponding bounding boxes, respectively. 

The model is first trained on the given dataset $\mathcal{D}^A_{Train}$, to estimate $\theta^A$. Then, the model faces a new dataset $\mathcal{D}^B$ with a new distribution and makes the prediction of $Dets^{B|A} = \mathcal{P}(\theta^A,\Lambda,\mathcal{D}^B_{Test})$,
These predictions are likely to be less accurate than those of $Dets^B$. The goal of this work, therefore, is to develop a solution capable of generalizing $\mathcal{P}(\cdot)$
by tuning $\theta^{A}$ with $\mathcal{D}^B_{Train}$ so that the resulting detection model approximates both $Dets^A$ and $Dets^B$.
Let $\theta^{A\rightarrow B}$ be the network parameters first trained on $\mathcal{D}^A_{Train}$ and then tuned on $\mathcal{D}^B_{Train}$. The objective then is:
 \begin{equation}
Dets^I \sim \mathcal{P}(\theta^{A\rightarrow B},\Lambda,\mathcal{D}^I_{Test}), I = A\  \mbox{or}\  B
\end{equation}
This is a challenging task since the distribution of $\mathcal{D}^B$ is different from that of $\mathcal{D}^A$ and the model must learn both distributions together. 


In our proposed solution shown in Fig.~\ref{arch}, we employ Faster R-CNN~\cite{ren2015faster} as the backbone pedestrian detector. This network is one of the most powerful and popular architectures for object detection, especially in pedestrian-related domains~\cite{hasan2021generalizable}. Faster R-CNN consists of three blocks. First, the input image is passed to a convolutional feature extractor. 
Then, the feature maps are passed to a Region Proposal Network (RPN) where several regions of the image are identified as candidates for the presence of objects. These regions and feature maps are then passed to the final block, in which ROI pooling is applied. Finally, the binary class labels indicating the existence of a pedestrian or not are estimated, and the bounding boxes are calculated based on the anchor boxes.

To help the network learn the new distribution while maintaining its performance on the previous dataset, EWC is employed~\cite{kirkpatrick2017overcoming}. This algorithm minimizes the change in the model's output with respect to the initial dataset after fine-tuning on a new dataset. This method was proposed originally for classification and has been widely used for class incremental learning~\cite{liu2020incdet}. In this work, we modify EWC to address the problem of out-of-distribution pedestrian detection, which has a regression task as well.
We formulate the modified EWC loss as:
\begin{equation}
\mathcal{L}_{EWC}(\theta) = \sum_{i=1}^{k}\frac{\lambda}{2}\mathfrak{F}_i(\theta_i -\theta^A_{i})^2 ,
\end{equation}
where $\theta = \{\theta_i\}_{i=1}^k$ is a set of $k$ model parameters, $\lambda$ is a coefficient that controls the trade-off between learning the new distribution and remembering the previous one, $\theta^A$ is the network parameters after training on $\mathcal{D}^A_{Train}$, and $\mathfrak{F}_i$ encapsulates the importance of each weight. This parameter is calculated after training on the first dataset for each weight using:
\begin{equation}
\mathfrak{F}_i = \frac{1}{n}\sum_{X,Y \in \mathcal{D}^A_{Train}} \left ( \frac{\delta \log \mathcal{L}_A( \mathcal{P}(\theta,\Lambda,X),Y)}{\delta \theta_i} 
\left| \begin{matrix}
\\
\theta = \theta^A
\end{matrix}\right.  \right),
\end{equation}
where $n$ is the size of $\mathcal{D}^A_{Train}$, $X$ is an image from $\mathcal{D}^A_{Train}$, $Y$ is its corresponding ground truth, and $\mathcal{L}_{A}$ is the Faster R-CNN loss on this dataset.

The importance matrix in the original EWC loss is based on the model output ~\cite{knoedler2022improving, liu2020incdet}, whereas here we base our EWC loss on the network loss. This formulation has the advantage of allowing the two components of the network loss to be seamlessly integrated, regardless of differences between their absolute ranges. For example, class labels fall between zero and one whereas bounding box coordinates can vary based on the size of images and pedestrians. In our formulation, the importance of a parameter is now based on the derivative of the loss with respect to that parameter, rather than the derivative of the model's output which effectively balances the two loss quantities. 
Accordingly, our total loss for fine-tuning on $\mathcal{D}^B_{Train}$ after training on $\mathcal{D}^A_{Train}$ consists of two terms:
 \begin{equation}
 \mathcal{L}_{total} =  \mathcal{L}_{B}(\mathcal{P}(\theta,\Lambda,X),Y) +  \mathcal{L}_{EWC}(\theta).
\end{equation}
Here, $\mathcal{L}_{B}$ is the Faster R-CNN loss for $\mathcal{D}^B_{Train}$ which is is the sum of classification loss ($\mathcal{L}_{cls}$) and bounding box regression loss ($\mathcal{L}_{reg}$). Note that our modified EWC loss can also be applied to pure classification, as well as pure regression tasks. Another strength of EWC compared to other CL methods is that $\lambda$ provides an ability to easily control the relative rate of learning vs. forgetting. This is a particularly attractive property in pedestrian detection, as outdoor locations exhibit changes to ambient conditions at different rates.




Previous studies point to a potential disadvantage of using EWC for training, in that it may result in 
exploding gradients~\cite{liu2020incdet}. In order to overcome this issue with minimum modification to the loss and maintain its advantages, all gradients are clipped to $G_{\max}$ in the training process, with values specified in Sec.~\ref{ES}. 
Our generalization of $\mathcal{P}(\cdot)$ in Eq. \ref{eq:1} is addressed by using $\mathcal{L}_{total}$ which includes $\mathcal{L}_{EWC}(\theta)$, to effectively learn $\theta^{A\rightarrow B}$.

\section{Experiments and Results}
\subsection{Experimental Setup}\label{ES}
\noindent \textbf{Datasets.} We use two popular datasets for human detection, CityPersons~\cite{zhang2017citypersons} and CrowdHuman~\cite{shao2018crowdhuman}. Both datasets have been used in many recent studies and serve as standard benchmarks for performance evaluation of pedestrian detection models~\cite{hasan2021generalizable}. They contain 5k and 15k images, respectively.
We have used CityPersons and CrowdHuman given
their considerable distribution shifts. To illustrate this we
present a UMAP of several common datasets in this area, which shows our two selected datasets are
further apart compared to others commonly used in the field.

\begin{figure}[t]
\begin{center}
\includegraphics[width = 0.8\linewidth]{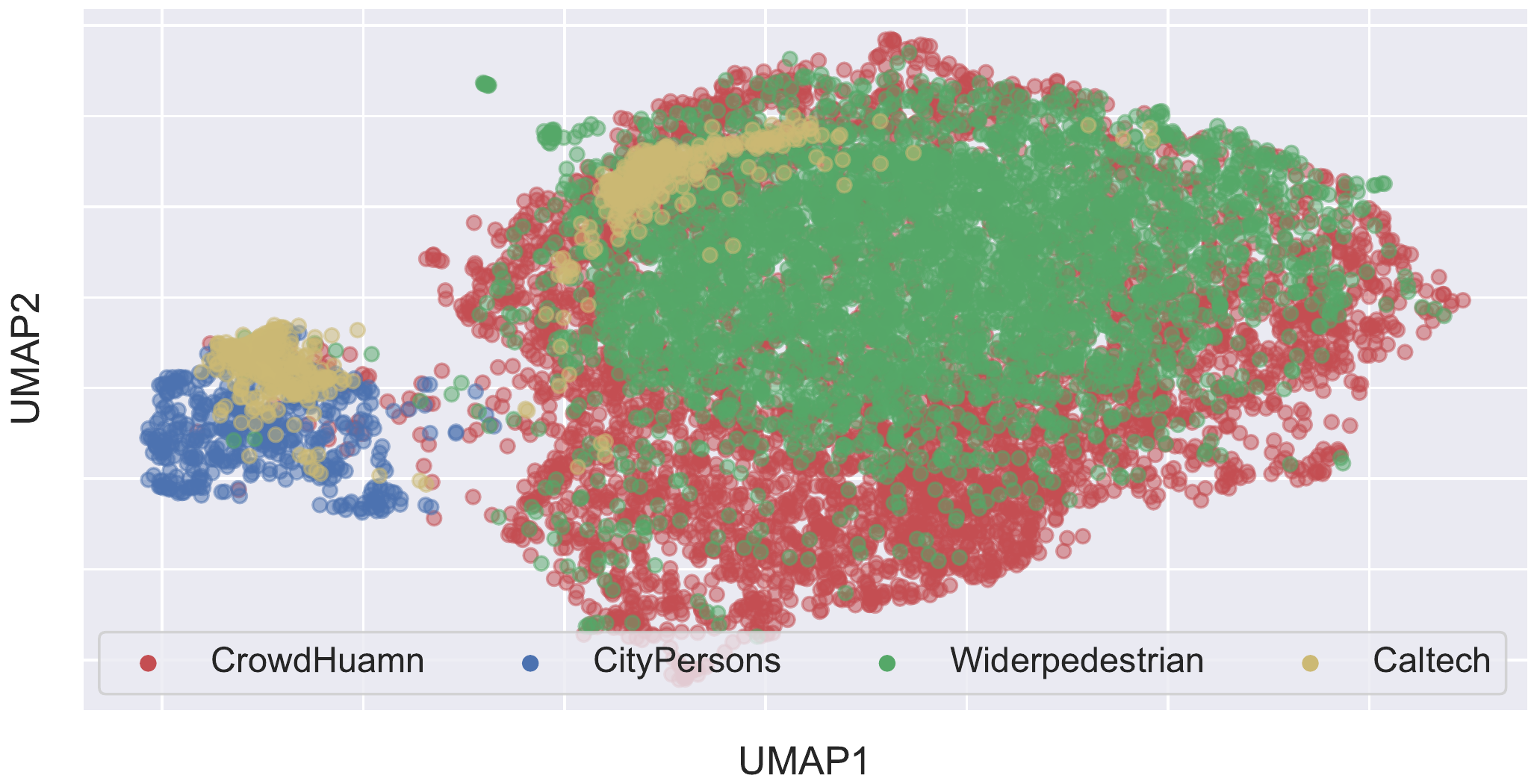}
\end{center}
\caption{The distribution of some of the benchmark datasets in the field of human detection.}
\label{fig:dist}
\end{figure}


\noindent \textbf{Implementation Details.} In this work, we utilize ResNet-50 with pre-trained ImageNet weights as the feature extractor of Faster R-CNN~\cite{he2016deep}. SGD optimizer was used with an initial learning rate of 5E-3, a momentum of 0.9, and a Weight decay of 5E-4. The model was trained for 10 epochs on each dataset. $G_{\max}$ was set to 20 and $\lambda$ to 1E-6 for CrowdHuman and 1.2E-6 for CityPersons. The optimum value of $\lambda$ is selected using a grid search. All training was performed with an NVIDIA RTX 3090 GPU. Upon publication, we will make our implementation publicly available at: \href{https://github.com/MahdiyarMM/Continual-pedestrian-detection}{link}.

\noindent \textbf{Evaluation.} The widely accepted metric for pedestrian detection is log miss rate average over False Positive Per Image (FPPI),
denoted as $MR^{-2}$~\cite{hasan2021generalizable, zhang2017citypersons}. 
$MR^{-2}$ is calculated following the same setup described in ~\cite{hasan2021generalizable}, so only those pedestrian bounding boxes with a height greater than 50 pixels are considered in the calculation. Its value is reported based on varying occlusion ratios, `Reasonable', `Bare', `Partial', and `Heavy',  which represent the occlusion ratios of [0, 0.35], 0, (0, 0.35], and (0.35,0.8], respectively.

\subsection{Results}The model is first trained on $\mathcal{D}^A_{Train}$,
and then fine-tuned in two different ways on $\mathcal{D}^B_{Train}$, once without any regularization loss and once using our proposed loss. The Reasonable, Bare, Partial, and Heavy $MR^{-2}$ of both trained models on both test sets are evaluated. Table~\ref{res_i} presents results for two scenarios. In the first scenario, $\mathcal{D}^A=$ CrowdHuman and $\mathcal{D}^B=$ CityPersons, and vice versa in the second scenario. In this table, we observe that after baseline fine-tuning on the second dataset, the $MR^{-2}$ of the first dataset increases (i.e., degrades) notably, from 55.48 to 63.20 for CrowdHuman and from 23.59 to 35.96 for CityPersons. This is due to the difference in the distributions of the two datasets, and the fact that recent deep learning models exhibit challenges when generalizing from one dataset to another in pedestrian detection tasks~\cite{hasan2021generalizable}. However, when the model is fine-tuned using our proposed approach, the performance drop is considerably improved (3.01 vs. 7.72 for CrowdHuman and 8.07 vs. 12.37 for CityPersons in Reasonable $MR^{-2}$). Moreover, after fine-tuning, the model reaches even better performance on the first dataset than when training from scratch, indicating that by applying the proposed method,
useful representations from the second dataset boost performance on the first dataset. 
This experiment also serves as an ablation study, since the impact of our loss is compared to the vanilla model.

\begin{figure}[t]
    \begin{subfigure}[b]{0.48\linewidth}
        \centering
        \includegraphics[width = 1.1\linewidth]{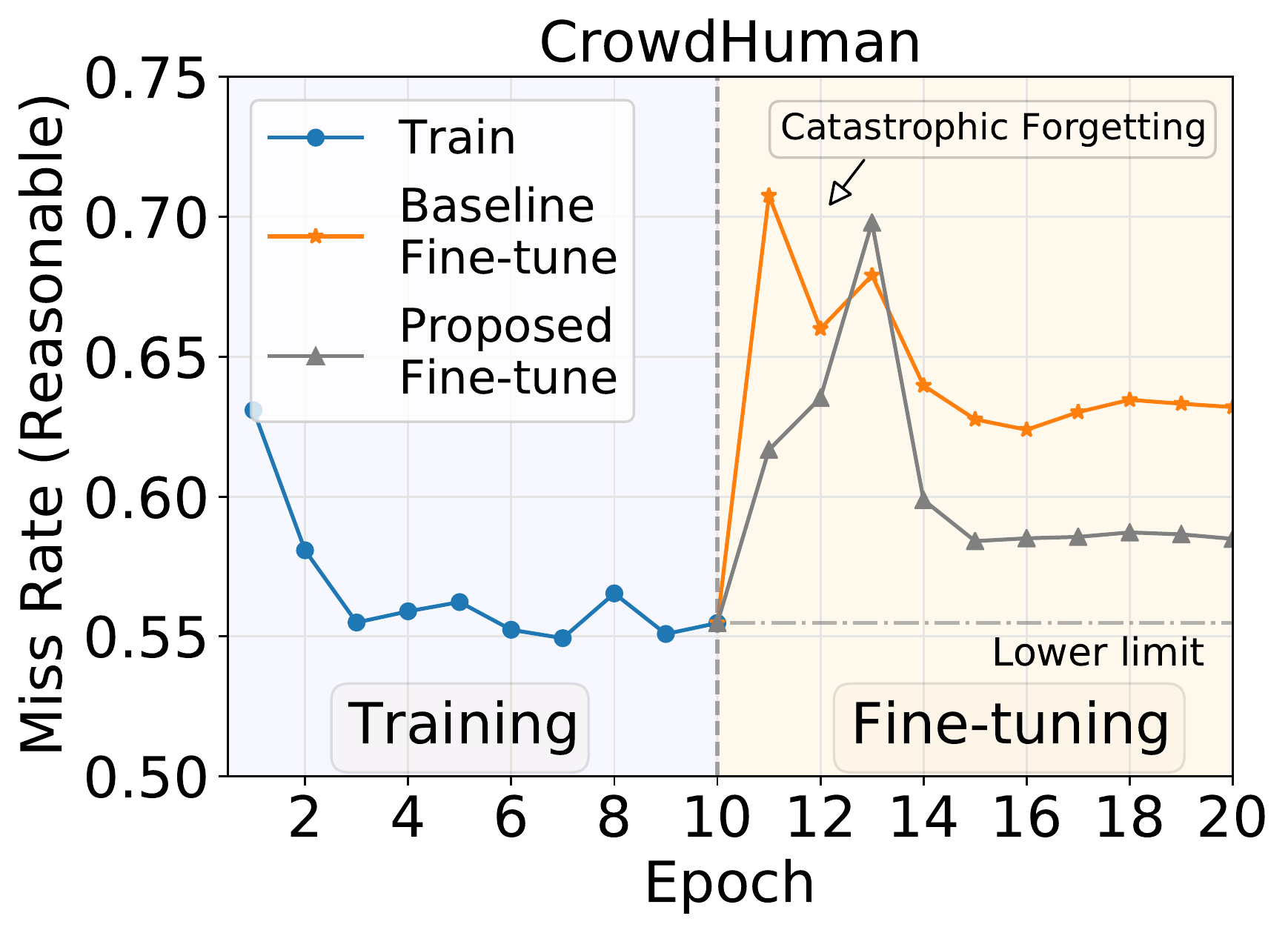}
    \end{subfigure}
    \hfill
    \begin{subfigure}[b]{0.48\linewidth}
        \centering
        \includegraphics[width = 1.1\linewidth]{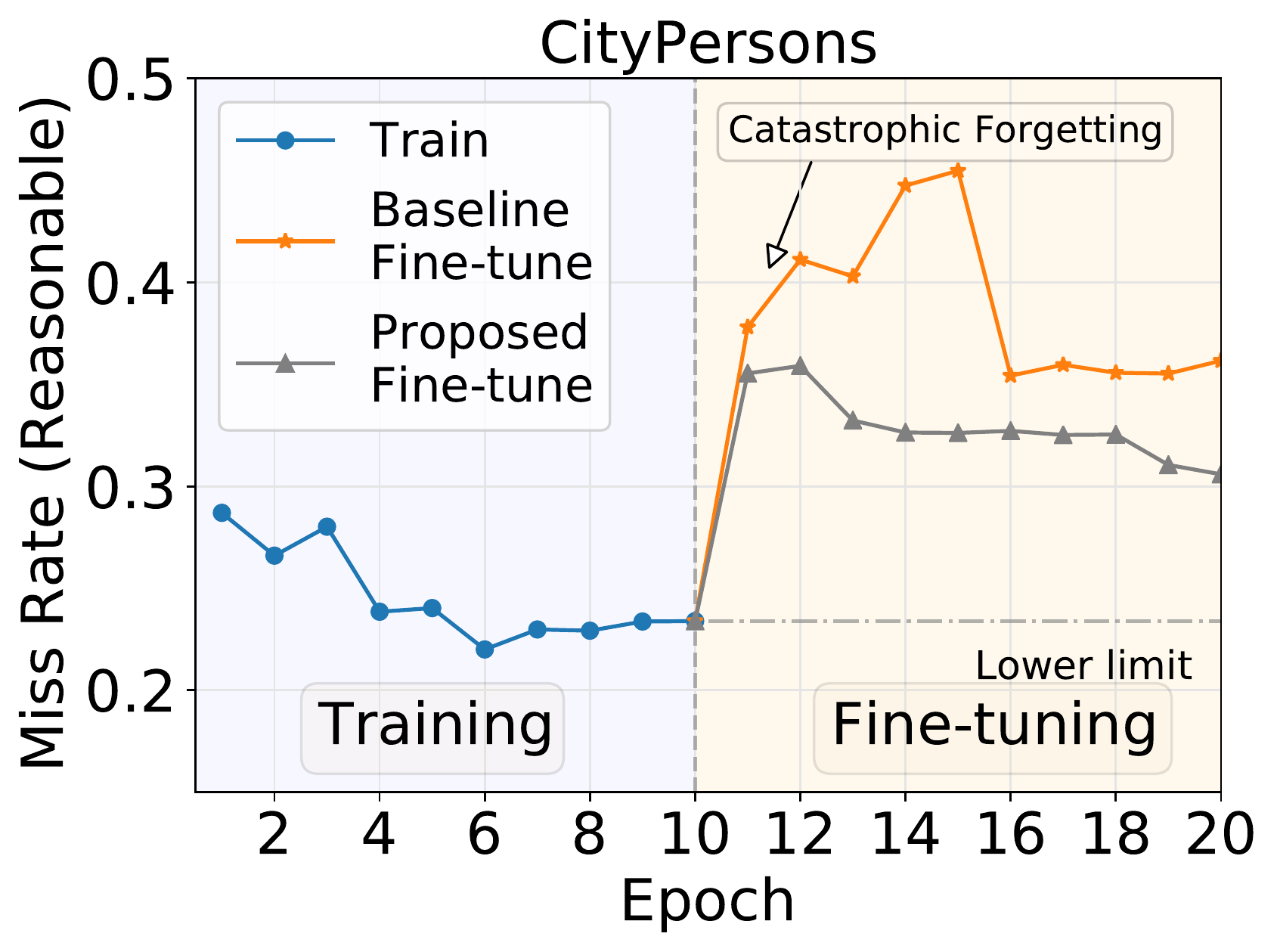}
    \end{subfigure}
    \centering
\caption{Miss rate of (left) scenario 1: CrowdHuman and (right) scenario 2: CityPersons for train and fine-tune phase.}
\label{fig:res}
\end{figure}


The performance of the model is plotted against epochs in both training and fine-tuning phases in Fig.~\ref{fig:res}. The increase in $MR^{-2}$ at epoch 10 indicates catastrophic forgetting, which occurs when a model learns a new task or data distribution. Comparing the baseline against our proposed solution illustrates our modified EWC helps the model learn the new data with a minimal drop in the performance on the previous data. 

To compare the performance of the proposed model with other architectures for out-of-distribution pedestrian detection, the percentage increase of $MR^{-2}$ after fine-tuning on the second dataset is presented in Table~\ref{new_comp}. We observe that our proposed solution limits the $MR^{-2}$ increase notably in comparison to other pedestrian detection models. 

Finally, to further investigate the impact of the proposed model in different occlusion levels, the improvement of the $MR^{-2}$ in each group is shown in Table.~\ref{new_comp_3}. We observe consistent improvement for both datasets over different occlusion rates, indicating that the improvement is not limited to a particular occlusion type/rate. We conclude that our proposed model is a step towards a general pedestrian detection network that can perform well on different distributions.

\begin{table}[t!]
\footnotesize
\centering
\caption{Miss rate ($MR^{-2}(\downarrow)$)  percentage increase.} 
\setlength
\tabcolsep{3pt}
\begin{tabular}{l|c|c}
\hline
\multirow{2}{*}{\textbf{Model}}       & \textbf{Scenario 1} & \textbf{Scenario 2}  \\
                                      & \textbf{CrowdHuman} & \textbf{CityPersons} \\ \hline
Faster R-CNN (ResNet-no TL) \cite{zhang2017citypersons} & 30\%                & 41\%                 \\
Faster R-CNN (MobileNet) \cite{zhang2017citypersons}               & 13\%                & 62\%                 \\
Faster R-CNN (ResNet) \cite{zhang2017citypersons}                  & 14\%                & 54\%                 \\
FPN \cite{lin2017feature}                                   & 16\%                & 43\%                 \\
Cascade R-CNN \cite{cai2019cascade}                         & 17\%                & 41\%                 \\
HRNet \cite{wang2020deep}                                & -                  & 40\%               \\
Swin-Trans \cite{liu2021swin}                           & -                   & 81\%                 \\
Proposed                              & \textbf{5\%}                 & \textbf{34\%}                \\ \hline
\end{tabular}
\label{new_comp}

\end{table}

\begin{table}[t!]
\footnotesize
\centering
\caption{Miss rate improvement for both datasets for different occlusion rates.} 
\setlength
\tabcolsep{3pt}
\begin{tabular}{l|cccc}
\hline
\multirow{2}{*}{\textbf{Dataset}} & \multicolumn{4}{c}{\textbf{Occlusion Rate}}                             \\
                                  & \textbf{Reasonable} & \textbf{Bare} & \textbf{Partial} & \textbf{Heavy} \\ \hline
CityPersons                       & 4.71                & 5.34          & 6.49             & 4.15           \\
CrowdHuman                        & 4.54                & 3.46          & 3.51             & 4.33           \\ \hline
\end{tabular}
\label{new_comp_3}

\end{table}

\section{Conclusion}
We have used continual learning to address out-of-distribution generalization for pedestrian detection. Recent pedestrian detection models are heavily biased toward their training distributions, and thus fine-tuning them on a new distribution leads to catastrophic forgetting. Here, we propose a new solution by modifying Elastic Weight Consolidation (EWC) and applying it to a standard pedestrian detection network. Our detailed experiments show that the model effectively learns the new distribution with minimum forgetting of the previous data. 
In future work, a replay memory will be used in fine-tuning alongside our proposed loss to further improve effectiveness. Moreover, vision transformers will be used along with self-supervised pre-training.


\noindent \textbf{Acknowledgements.} We would like to thank Geotab Inc., the City of Kingston, and NSERC for their support of this work.


\bibliographystyle{IEEEbib}
\small
\bibliography{refs}

\end{document}